\documentclass[authoryear,5p]{elsarticle}
\usepackage[export]{adjustbox}
\usepackage{lineno,hyperref}
\usepackage{dirtytalk}
\usepackage{tikz}
\usepackage{tikz-qtree}
\usepackage{color}
\usepackage{mathtools}
\usepackage{comment}
\usepackage{rotating}
\usepackage{cancel}
\usepackage{algorithm}
\usepackage{algpseudocode}
\usepackage{multirow}
\usepackage{multicol}
\usepackage{graphicx}
\usepackage{booktabs}
\usepackage{listings}
\usepackage{soul}
\usepackage[multiple]{footmisc}
\usepackage[skip=0.2\baselineskip]{caption}
\usepackage{mathtools}

\usepackage{soul,color}
\soulregister\cite7

\makeatletter
\newcommand\footnoteref[1]{\protected@xdef\@thefnmark{\ref{#1}}\@footnotemark}
\makeatother

\modulolinenumbers[5]

\journal{}
\biboptions{authoryear}

\begin{document}

\lstset{basicstyle=\normalsize\ttfamily,breaklines=true}

\begin{frontmatter}

\title{Detection of Temporality at Discourse Level on Financial News by Combining Natural Language Processing and Machine Learning}

\author[mymainaddress]{Silvia Garc\'ia-M\'endez\corref{mycorrespondingauthor}}
\ead{sgarcia@gti.uvigo.es}
\author[mymainaddress]{Francisco de Arriba-P\'erez}
\ead{farriba@gti.uvigo.es}
\author[mymainaddress]{Ana Barros-Vila}
\ead{abarros@gti.uvigo.es}
\author[mymainaddress]{Francisco J. Gonz\'alez-Casta\~no}
\ead{javier@det.uvigo.es}
\address[mymainaddress]{Information Technologies Group, atlanTTic, University of Vigo, E.I. Telecomunicaci\'on, Campus, 36310 Vigo, Spain}

\cortext[mycorrespondingauthor]{Corresponding author: sgarcia@gti.uvigo.es}

\begin{abstract}

Finance-related news such as Bloomberg News, CNN Business and Forbes are valuable sources of real data for market screening systems. In news, an expert shares opinions beyond plain technical analyses that include context such as political, sociological and cultural factors. In the same text, the expert often discusses the performance of different assets. Some key statements are mere descriptions of past events while others are predictions. Therefore, understanding the temporality of the key statements in a text is essential to separate context information from valuable predictions. We propose a novel system to detect the temporality of finance-related news at discourse level that combines Natural Language Processing and Machine Learning techniques, and exploits sophisticated features such as syntactic and semantic dependencies. More specifically, we seek to extract the dominant tenses of the main statements, which may be either explicit or implicit. We have tested our system on a labelled dataset of finance-related news annotated by researchers with knowledge in the field. Experimental results reveal a high detection precision compared to an alternative rule-based baseline approach. Ultimately, this research contributes to the state-of-the-art of market screening by identifying predictive knowledge for financial decision making.

\end{abstract}

\begin{keyword}
Computational Linguistics; financial news; knowledge extraction; Machine Learning; Natural Language Processing; temporal analysis.
\end{keyword}

\end{frontmatter}

\section{Introduction}

The field of Computational Linguistics has been prolific in theoretical work and practical solutions to real world challenges. This has become possible thanks to the increase in computing power, the development of enhanced architectures and Machine Learning models, and the advent of Web 4.0 online sources of relevant information. Natural Language Processing ({\sc nlp}) is being broadly applied to these sources, from simple solutions using for example part-of-speech ({\sc pos}) data, to more sophisticated designs exploiting syntactic and semantic dependencies. 

Artificial understanding of a text to describe its meaning at discourse level, such as its temporal dimension, remains an open research question. Current work on knowledge extraction from text based on Artificial Intelligence ({\sc ai}) techniques only covers limited aspects \citep{Collobert2011}.

Finance-related news from sources such as Bloomberg News, CNN Business and Forbes are valuable real data for market screening systems. News describe not only technical analyses of assets' performance but also their political, sociological and cultural contexts. Their content is informally organised around key statements that carry the main opinions of the author. Thus, the analysis of the temporal dimension of key statements becomes essential to differentiate predictions from the rest of the text, which also provides valuable information to interpret these statements, including their temporality.

In fact, every sentence in a text holds temporal knowledge \citep{Zwaan1996}. Temporal representation is an innate human capacity related to cognition and discourse processing, and there exist plenty of linguistic elements to express it \citep{Demagny2012}. More specifically, it involves certain linguistic markers \citep{Evers-Vermeul2017} that typically combine lexical (temporal meaning of words), morphological (temporal features of languages such as tense), syntactic (position of time markers within the clause and their relation to other constituents) and pragmatic (discourse organisation and coherence through temporality) processings. Thus, temporal markers are not simply grammatical categories related to features like tense and aspect, but also temporal adverbial elements (for example: ``a month ago'', ``today'', ``after'', etc.) and complex verb structures such as phrasal verbs or compound verb phrases (e.g., ``look forward'', ``begin to work'').

Even though human beings excel in associative discourse inference, transferring this knowledge to computational systems is not straightforward \citep{Pratt2001,Pratt-Hartmann2005}. In this work, we address temporal analysis at discourse level with three different strategies. First, we apply clause segmentation to determine the continuity of tense in sequential clauses within the text, both across dependencies and by proximity. Second, we detect temporal modifiers, more specifically expressions that directly modify the temporal data of a clause, for example by altering verb tense. Finally, the positions of temporal references within the news are also considered, since authors tend to follow certain patterns, such as leaving predictions (future tenses) to the end. We apply these strategies along with Machine Learning techniques to determine the temporality of key opinions about assets in news.
To the best of our knowledge, this work represents the first attempt to perform temporal analysis of finance news at discourse level. The rest of this article is organised as follows: Section \ref{sec:related_work} reviews related work on knowledge extraction and temporal analysis at discourse level in economics. Section~\ref{sec:system} describes the model of the problem and our solution. Section \ref{sec:results} presents the experimental text corpus and the numerical tests that validate our approach. Finally, Section \ref{sec:conclusions} concludes the paper.

\section{Related work}
\label{sec:related_work}

Stock market screening with Data Mining has produced a significant body of research with successful outcomes that effectively support investment decision making \citep{Alanyali2013,Day2016}. Works such as \cite{Karabulut2012,Nofer2015,Dimpfl2016} have demonstrated the strong correlation between volatility and volume of search queries about stock market indexes.

In this field, {\sc nlp} techniques have been successfully applied to noise removal and feature extraction \citep{Sun2014,liu2015sentiment,Fisher2016,Xing2018} from financial reports such as news \citep{Zhang2010,Alanyali2013,Atkins2018}, micro-blogging comments \citep{Sun2014,Fisher2016,Rickett2016,Wang2017,Xing2018} and social media \citep{Ioanas2014,Sun2016}. These techniques have been often combined with Machine Learning algorithms \citep{Huang2012,Prollochs2015}, which can be divided into supervised (based on manual annotations) \citep{Alanyali2013,Prollochs2015} and unsupervised approaches \citep{Huang2012,Prollochs2015}.

\cite{Alanyali2013} demonstrated the relation between stock market events and public data in financial news. \cite{Atkins2018} applied Machine Learning models of Latent Dirichlet Allocation to predict stock market volatility. They concluded that the information extracted from news sources was more adequate than price variations for predicting the volatility of financial assets. Other authors have applied Deep Learning algorithms to financial news \citep{Day2016,Vargas2017}. Specifically, \cite{Day2016} concluded that the particular typology of the source (news, micro-blogging comments, etc.) has strong influence in the resulting decision; while \cite{Vargas2017} proved that recurrent Neural Networks are better at capturing context information and modelling complex temporal characteristics for stock market forecasting.

Despite the several works on stock market knowledge extraction, the analysis of the temporal dimension deserves attention, since in most of the literature \citep{Zhang2010,Rickett2016,Sun2016,Wang2017,Atkins2018}, temporality is exclusively determined from news, posts or comment timestamps. Early research \citep{Schumann1987} examined the expression of temporality from different linguistic perspectives: morphology, semantics and pragmatics. Strongly aligned with our view, \cite{Gibbs1998} highlighted the importance of analysing the time dimension from the customer perspective, i.e. the consumers’ own understanding of time. Furthermore, \cite{Forray2005} extracted time-related knowledge from journal titles. They paid attention to features such as punctuation, word choice, the use of academic terminology and keywords, as well as to context information. 

Among recent research, there exist three general approaches to temporal analysis. First, in \citep{Kehler2002}, temporal relations are simply listed among other discourse relations. Conversely, in other works, different types of temporal relations constitute a separate class of discourse relation. This is the case of the Penn Discourse Treebank \citep{Prasad2008} and the Rhetorical Structure Theory Discourse Treebank \citep{Carlson2001}. Finally, the third approach does not consider temporal order as a relational feature but as a pragmatic segment-specific feature \citep{Sanders1993}.

Considerable research on temporality has focused on verbs and, more specifically, on the analysis of verb tenses \citep{Karapandza2016,Evers-Vermeul2017}. Tense is a language feature that situates an event in time \citep{Salaberry2003}. Thus, it allows to contextualise the event within a temporal frame. In the specific case of English, it follows a rather clear scheme that simplifies the comprehension of temporal discourse organisation \citep{Demagny2012}. In other words, past, present and future tense morphologies often represent the corresponding times accurately.

In this paper, we also employ temporality at sentence level (the location of events on a timeline as indicated by past, present or future tenses), but we focus on the discourse level, that is, the relational organisation of the discourse due to the temporal ordering. As far as we know, no other state-of-the-art research has analysed the temporality of financial texts as expressed in natural language in the content itself.

\section{System architecture}
\label{sec:system}

\begin{figure*}[ht!]
\centering
\includegraphics[scale=0.23]{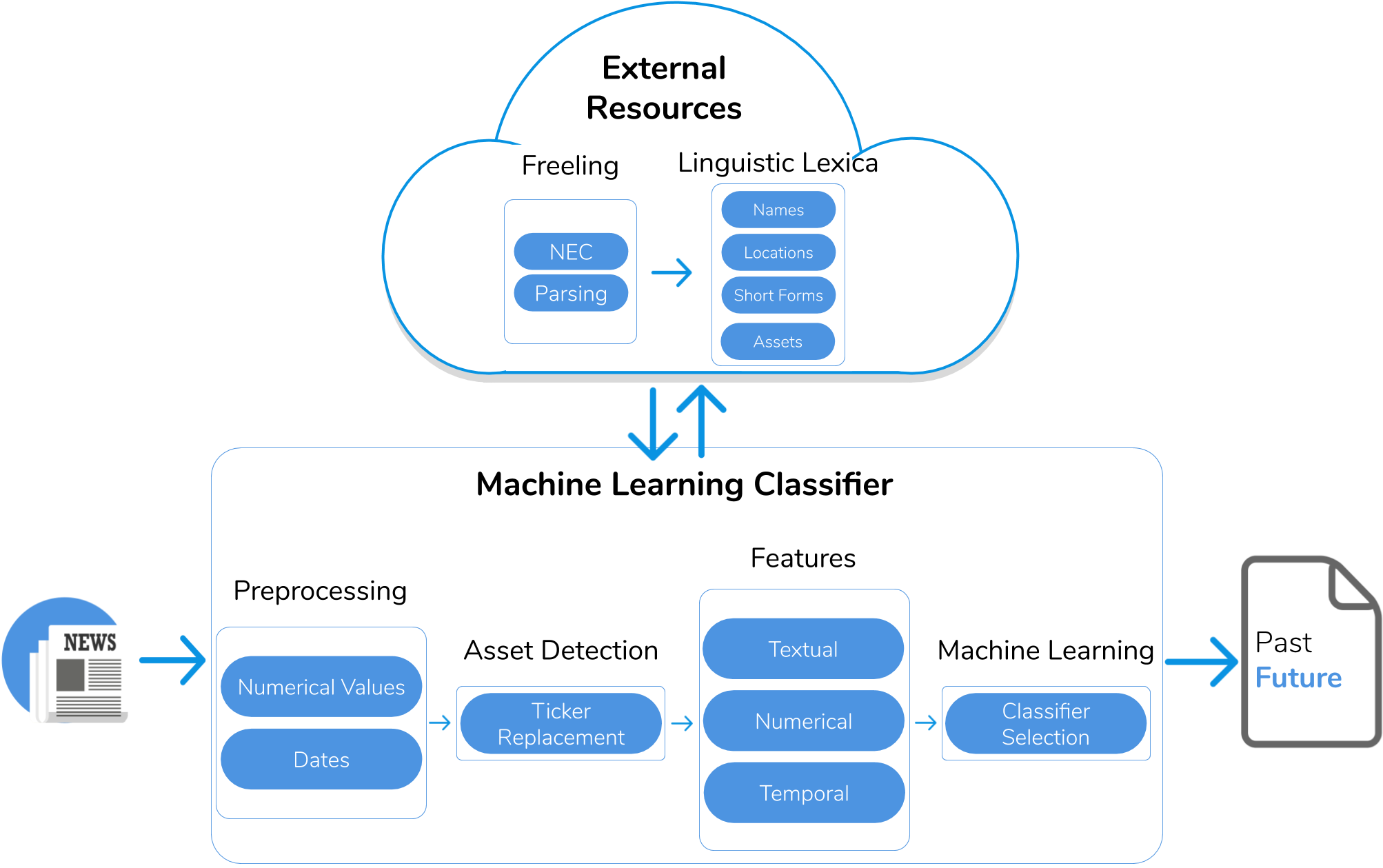}
\caption{\label{architecture} System architecture.}
\end{figure*}

In this section we present a novel two-stage system to detect at discourse level the temporality of finance-related news, by combining {\sc nlp} techniques, to extract sophisticated features like syntactic and semantic implicit dependencies, with a Machine Learning model. We seek to detect the general tense (past or future) when the author expresses an opinion about a stock asset, but we are specially interested in the future, which is strongly related to predictive analysis. To determine syntactic and semantic dependencies, we consider explicit mentions to stock market assets acting as subjects and objects in the sentences. Figure \ref{architecture} shows the general architecture of our system. In the next subsections we describe its modules.

\subsection{Preprocessing}
\label{prepro}

This module comprises the different techniques to remove unnecessary and redundant information from the input dataset. 

First, before the replacement procedures, it is necessary to homogenise numerical data and dates in the experimental dataset. We rewrite them to a common format by applying regular expressions. Next, we replace all numerical values including percentages and dates with $NUM$, $PERC$ and $DATE$ tags, respectively.

Then, we use the Name Entity Classification ({\sc nec}) functionality from Freeling \citep{Atserias06,PadroEtAl12} to detect proper names and locations. We double-check this detection with freely available lexica\footnote{Available at {\tt https://names.mongabay.com/most\_common\_sur-
names.htm}, November 2020.}\footnote{Available at {\tt https://datahub.io/core/world-cities\#readme}, November 2020.}. We also use freely available lexica to detect abbreviations\footnote{Available at {\tt https://www.gti.uvigo.es/index.php/en/resour-
ces/9-abbreviation-lexicon}, November 2020.}. At the end, these elements are replaced by tags $NAME$ (proper names), $LOC$ (locations), and $ABB$ (abbreviations).

\subsection{Asset detection}
\label{asset}

Notwithstanding the difficulty to extract stock market assets when neither exact names nor pronouns are used, we detect referential elements, such as nouns (e.g. ``company'', ``enterprise'', ``stock'', ``investment'', ``opportunity'', ``share'', ``manufacturer'', etc.). More specifically, if the referential element appears after one or several tickers (within the same sentence), it is replaced by the last ticker; otherwise, it is replaced by the last ticker of the previous sentence (if any).

We use tag $TICKER$ to replace the main asset in the news, as indicated by the annotators, and tag $OTHER$ for any other assets. All these other assets in each news piece are extracted using a finance lexicon\footnote{Available at {\tt https://www.gti.uvigo.es/index.php/en/resour-
ces/10-financial-lexicon}, November 2020.}. Table \ref{tab:beforeafterpreproOriginal} shows a complete example of asset detection and replacement (note that it also exemplifies numerical data and date detection and replacement, see Section \ref{prepro}).

\begin{table*}[!htbp]
\centering
\caption{Example of news content before and after processing.}
\begin{tabular}{cc}
\hline
& \textbf{News content} \\ \toprule
Before & \begin{tabular}[c]{@{}c@{}}If they could get the planes, {\bf Airbus} and {\bf Boeing} are sold out through {\bf 2023}. \\ {\bf On October 29}, {\bf 2018}, the {\bf stock} dropped {\bf 6.6\%} and recovered in three days.
\end{tabular}
\\ \hline
After & \begin{tabular}[c]{@{}c@{}}If they could get the planes, {\bf OTHER} and {\bf TICKER} are sold out through {\bf DATE}. \\ On {\bf DATE}, {\bf DATE}, the {\bf TICKER} dropped {\bf NUM} and recovered in three days.
\\
\end{tabular}
\\ \bottomrule
\end{tabular}
\label{tab:beforeafterpreproOriginal}
\end{table*}

\subsection{Features and Machine Learning analysis}
\label{features}

Our system employs ZeroR, Decision Tree ({\sc dt}), Random Forest ({\sc rf}), Linear Support Vector Classification ({\sc svc}) and Neural Network ({\sc nn}) algorithm implementations from the Scikit-Learn Python library\footnote{Available at {\tt https://scikit-learn.org/stable/supervised\_
learning.html\#supervised-learning}, November 2020.}. We included ZeroR as a reference for the performance of the rule-based baseline classifier due to its simplicity and low computational cost \citep{Fazayeli2019}.

Table {\ref{tab:complexity}} shows the training and testing complexity of the Machine Learning algorithms we used in our analysis for $c$ target classes, $f$ features, $i$ algorithm instances (where applicable) and $s$ dataset samples. For the specific case of the {\sc nn} algorithm, $m$ represents the number of neurons, and $l$ its layers. The {\sc{dt}} and {\sc{rf}} algorithms have logarithmic training complexity \citep{Witten2016,Hassine2019}, that is, less than {\sc{svc}} \citep{Vapnik2000} (which, however, has a low classification response time compared to other alternatives when using a linear kernel, as in our work). At the end, {\sc nn} has the highest training and testing complexity \citep{Han2012,Witten2016}.

\begin{table}[ht!]
\caption{\label{tab:complexity} Machine Learning training and testing complexity order.}
\centering
\begin{tabular}{lll}\bottomrule
\multicolumn{1}{c}{\bf Classifier} & \multicolumn{1}{c}{\bf Train complexity} & \multicolumn{1}{c}{\bf Test complexity}\\\hline
DT & O($n \cdot \mbox{log}(n) \cdot f$) & O(depth of the tree)\\
RF & O($n\cdot \mbox{log}(n) \cdot f \cdot i$) & O(depth of the tree$\cdot i$)\\
SVC & O($s^2$) & O($f$)\\
NN & O($m \cdot f \cdot s \cdot l$) & O($m \cdot f \cdot l$)
\\\toprule
\end{tabular}
\end{table}

\begin{figure*}[ht!]
\centering
\includegraphics[scale=0.35]{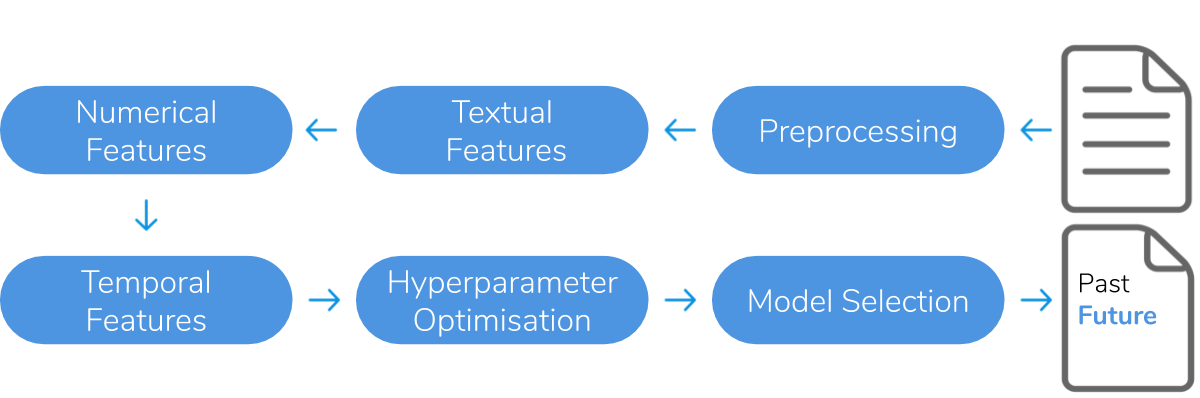}
\caption{\label{flow_diagram_ML}Flow diagram of Machine Learning analysis.}
\end{figure*}

Figure \ref{flow_diagram_ML} shows the Machine Learning process. There are $f=$30 features in our system as indicated in Table \ref{tab:features}, divided into textual, numerical and temporal features. 

\begin{table*}[ht!]
\centering
\small
\caption{\label{tab:features} Features of the temporality detection system.}
\begin{tabular}{lll}
\hline
\bf Type & \bf Feature name & \bf Description \\\toprule

Textual & CHAR\_GRAMS & \begin{tabular}[c]{@{}p{9.5cm}@{}} Character $n$-grams from the news content.\end{tabular}\\

 & WORD\_TOKENS & \begin{tabular}[c]{@{}p{9.5cm}@{}} Character $n$-grams only from text inside word boundaries.\end{tabular}\\

 & WORD\_GRAMS & \begin{tabular}[c]{@{}p{9.5cm}@{}} Word $n$-grams from the news content.\end{tabular}\\ \hline

Numerical & NUM & \begin{tabular}[c]{@{}p{9.5cm}@{}} Amount of numerical values (excluding percentages) in the news content.\end{tabular}\\

 & PERC & \begin{tabular}[c]{@{}p{9.5cm}@{}} Amount of percentages in the news content.\end{tabular}\\ \hline

Temporal & PRS\_DEP\_SUB & \begin{tabular}[c]{@{}p{9.5cm}@{}} Amount of verbs in present tense from the dependency analysis when the ticker acts as subject.\end{tabular}\\

 & PST\_DEP\_SUB & \begin{tabular}[c]{@{}p{9.5cm}@{}} Amount of verbs in past tense from the dependency analysis when the ticker acts as subject.\end{tabular}\\

 & FUT\_DEP\_SUB & \begin{tabular}[c]{@{}p{9.5cm}@{}} Amount of verbs in future tense from the dependency analysis when the ticker acts as subject.\end{tabular}\\

 & GLOBAL\_DEP\_SUB & \begin{tabular}[c]{@{}p{9.5cm}@{}} Global temporality by majority voting from the dependency analysis when the ticker acts as subject.\end{tabular}\\

 & PRS\_DEP\_SUB\_OBJ & \begin{tabular}[c]{@{}p{9.5cm}@{}} Amount of verbs in present tense from the dependency analysis when the ticker acts either as subject or object.
\end{tabular}\\

 & PST\_DEP\_SUB\_OBJ & \begin{tabular}[c]{@{}p{9.5cm}@{}} Amount of verbs in past tense from the dependency analysis when the ticker acts either as subject or object.\end{tabular}\\

 & FUT\_DEP\_SUB\_OBJ & \begin{tabular}[c]{@{}p{9.5cm}@{}} Amount of verbs in future tense from the dependency analysis when the ticker acts either as subject or object.\end{tabular}\\

 & GLOBAL\_DEP\_SUB\_OBJ & \begin{tabular}[c]{@{}p{9.5cm}@{}} Global temporality by majority voting from the dependency analysis when the ticker acts either as subject or object.\end{tabular}\\

 & PRS\_PROX\_SUB & \begin{tabular}[c]{@{}p{9.5cm}@{}} Amount of verbs in present tense from the proximity analysis when the ticker acts as subject.\end{tabular}\\

 & PST\_PROX\_SUB & \begin{tabular}[c]{@{}p{9.5cm}@{}} Amount of verbs in past tense from the proximity analysis when the ticker acts as subject.\end{tabular}\\

 & FUT\_PROX\_SUB & \begin{tabular}[c]{@{}p{9.5cm}@{}} Amount of verbs in future tense from the proximity analysis when the ticker acts as subject.\end{tabular}\\

 & GLOBAL\_PROX\_SUB & \begin{tabular}[c]{@{}p{9.5cm}@{}} Global temporality by majority voting from the proximity analysis when the ticker acts as subject.\end{tabular}\\

 & PRS\_PROX\_SUB\_OBJ & \begin{tabular}[c]{@{}p{9.5cm}@{}} Amount of verbs in present tense from the proximity analysis when the ticker acts either as subject or object.\end{tabular}\\

 & PST\_PROX\_SUB\_OBJ & \begin{tabular}[c]{@{}p{9.5cm}@{}} Amount of verbs in past tense from the proximity analysis when the ticker acts either as subject or object.\end{tabular}\\

 & FUT\_PROX\_SUB\_OBJ & \begin{tabular}[c]{@{}p{9.5cm}@{}} Amount of verbs in future tense from the proximity analysis when the ticker acts either as subject or object.\end{tabular}\\

 & GLOBAL\_PROX\_SUB\_OBJ & \begin{tabular}[c]{@{}p{9.5cm}@{}} Global temporality by majority voting from the proximity analysis when the ticker acts either as subject or object.\end{tabular}\\

 & PRS\_INITIAL & \begin{tabular}[c]{@{}p{9.5cm}@{}} Amount of verbs in present tense in the initial third of the news.\end{tabular}\\

 & PST\_INITIAL & \begin{tabular}[c]{@{}p{9.5cm}@{}} Amount of verbs in past tense in the initial third of the news.\end{tabular}\\

 & FUT\_INITIAL & \begin{tabular}[c]{@{}p{9.5cm}@{}} Amount of verbs in future tense in the initial third of the news.\end{tabular}\\

 & PRS\_MEDIUM & \begin{tabular}[c]{@{}p{9.5cm}@{}} Amount of verbs in present tense in the middle third of the news.\end{tabular}\\

 & PST\_MEDIUM & \begin{tabular}[c]{@{}p{9.5cm}@{}} Amount of verbs in past tense in the middle third of the news.\end{tabular}\\

 & FUT\_MEDIUM & \begin{tabular}[c]{@{}p{9.5cm}@{}} Amount of verbs in future tense in the middle third of the news.\end{tabular}\\

 & PRS\_FINAL & \begin{tabular}[c]{@{}p{9.5cm}@{}} Amount of verbs in present tense in the final third of the news.\end{tabular}\\

 & PST\_FINAL & \begin{tabular}[c]{@{}p{9.5cm}@{}} Amount of verbs in past tense in the final third of the news.\end{tabular}\\

 & FUT\_FINAL & \begin{tabular}[c]{@{}p{9.5cm}@{}}Amount of verbs in future tense in the final third of the news.\end{tabular}\\

\bottomrule
\end{tabular}
\end{table*}

Before computing the textual features, the text is preprocessed to replace capital by lowercase characters and remove punctuation marks (commas, colons, brackets, hyphens, exclamation and interrogation marks), accents and apostrophes, as well as characters $@$ and \#. This is applied to the textual content excluding the tags identified in prior preprocessing and asset detection stages (sections {\ref{prepro}} and {\ref{asset}}, respectively).

At the end, the resulting textual features are $n$-grams. More specifically, char-grams, word tokens (character $n$-grams only from text inside word boundaries) and word-grams. We explain the $n$-gram feature selection in Section \ref{tuning_training}.

The numerical features are basically the numerical values in the news content. There are two such features: the amount of numerical values excluding percentages and the amount of percentages in the text. Note that we discard explicit dates, although it would be possible to determine if such dates refer to the future or the past if the news publication date is available. 

Regarding temporal features, we use the dependency parsing by Freeling to extract the verbs related to the tickers and their tenses. In addition, we analyse the proximity between a verb and a ticker within a clause in forward and backward mode. These analyses are applied regardless of the role of each ticker instance (subject or object). Taking the text {\em And second, will TICKER be allowed to execute its plan in full? Cost-cutting should make TICKER a leaner[...]} as an example, note that the first $TICKER$ tag acts as a subject while Freeling considers the second tag an object.

Observe that, in the dependency analysis, in case the ticker is a modifier of the subject (e.g. ``Intel stock'', where ``stock'' is the subject and ``Intel'' its modifier), and therefore it is not directly related to any verbs, we also consider the verbs that are linked to the subject. In the proximity analysis, given a ticker and the immediately preceding and posterior verbs within the clause, the nearest verb in number of intermediate words is selected. Table \ref{news_analysis} shows an example of news entry after applying dependency and proximity analysis. We highlight the tickers and the verbs that are related to them. Note that in both analyses the ticker may act as subject or object. Finally, we also consider the positions of the verbs in the discourse, by dividing the news pieces into three roughly equal parts in number of sentences. 

\begin{table*}[ht!]
\centering
\small
\caption{\label{news_analysis} Example of news entry after applying dependency and proximity analysis when the ticker acts as subject or object.}
\begin{tabular}{cc}
\hline
\bf Dependency analysis & \bf Proximity analysis \\\toprule
\begin{tabular}[c]{@{}p{7.5cm}@{}}
{\bf @TICKER@ is lagging} on its competitors. Make no mistake, @TICKER@ is going to have to fix @TICKER@ and @TICKER@ will take many, many, many years, @NAME@ Mosesmann, a technology analyst at @NAME@ Securities, told CNBC in an interview. Couple that issue with an ongoing search to replace former chief executive officer @NAME@ Krzanich, and {\bf @TICKER@ has} a lot on its plate heading into the last quarter of @DATE@ and beyond.
\end{tabular} & 
\begin{tabular}[c]{@{}p{7.5cm}@{}}
{\bf @TICKER@ is lagging} on its competitors. Make no mistake, {\bf @TICKER@ is going to have to fix} @TICKER@ and {\bf @TICKER@ will take} many, many, many years, @NAME@ Mosesmann, a technology analyst at @NAME@ Securities, told CNBC in an interview. Couple that issue with an ongoing search to replace former chief executive officer @NAME@ Krzanich, and {\bf @TICKER@ has} a lot on its plate heading into the last quarter of @DATE@ and beyond.
\end{tabular}
\\\bottomrule
\end{tabular}
\end{table*}

Summing up, we propose three temporal features (number of past, present and future tenses, when the ticker acts as subject and when it acts either as subject or object) for the dependency analysis, the same for the proximity analysis, one feature with the global temporality for the dependency analysis and another similar feature for proximity analysis (both by majority voting). There are three additional features per discourse partition (initial, middle, final) that simply count the respective amounts of verb tenses. This completes 25 temporal features. 

In case of tie between past and future tenses, the latter prevails. Even though the system only predicts past and future temporality, it also takes present tenses as input. Note that we apply a hyperparameter optimisation to the set of classifiers. This allows us to select the best model based on its performance (see Section \ref{machinelearning_results}).

\section{Experimental results and discussion}
\label{sec:results}

In this section we first present some preliminary results using a rule-based approach as a baseline, and then evaluate our system. A rule-based reference is common practice in the literature when similar competitors are missing \citep{Araque2017,Cronin2017,Kim2020,SinghChauhan2020,Jorgensen2021}. All experiments were performed on a computer with the following hardware specifications:
\begin{itemize}
 \item Operating System: Ubuntu 18.04.2 LTS 64 bits
 \item Processor: Intel\@Core i9-9900K 3.60 GHz 
 \item RAM: 32GB DDR4 
 \item Disk: 500 Gb (7200 rpm SATA) + 256 GB SSD
\end{itemize}

\subsection{Dataset}
\label{dataset}

Our experimental dataset\footnote{The experimental dataset will be made available to other researchers on request.} is composed of $s=$600 news pieces with the following information: identifier, content, ticker (main stock market asset discussed), source and temporality (past and future tags of key opinions in the text). The sources of the news include prestigious journals such as The New York Times and stock messaging boards such as Bloomberg. Table \ref{news} shows an example of news entry in the dataset. Table \ref{tab:dataset_distribution} shows the distribution of its entries. The dataset is comparable to other datasets in previous research on knowledge extraction \citep{Li2018,Garcia2018}.

\begin{table*}[ht!]
\centering
\caption{\label{news} Example of news entry in the dataset.}
\begin{tabular}{cccc}
\hline
\bf Text & \bf Ticker & \bf Source & \bf Temporality \\\toprule
\begin{tabular}[c]{@{}p{7.5cm}@{}}
Pros and Cons to Buying Intel Stock\\
Intel is lagging on its competitors. Can its stock price holding up under the pressure? ``Make no mistake, Intel is going to have to fix this and it will take many, many, many years.", Hans Mosesmann, a technology analyst at Rosenblatt Securities, told CNBC in an interview. ``Their process technology disadvantage, which I think is broken, will take five, six, seven years. I don't think that business model works by them being behind by a year or two in terms of process technology."
Couple that issue with an ongoing search to replace former chief executive officer Brian Krzanich, and Intel has a lot on its plate heading into the last quarter of 2019 and beyond.
\end{tabular} & Intel & US News & Future\\\bottomrule
\end{tabular}
\end{table*}

\begin{table*}[!htbp]
\centering
\caption{Distribution of entries in the dataset by category.}
\begin{tabular}{cccc}
\hline \textbf{Temporal tag} & \textbf{Number of entries} & {\bf Avg. length in sentences} & {\bf Avg. length in words} \\ \toprule
Past & 365 & 21.67 $\pm$ 13.89 & 351.57 $\pm$ 225.82\\
Future & 235 & 18.43 $\pm$ 12.41 & 303.40 $\pm$ 196.36\\\hline
Total & 600 & 20.05 $\pm$ 13.15 & 327.49 $\pm$ 211.09\\\bottomrule
\end{tabular}
\label{tab:dataset_distribution}
\end{table*}

\subsection{Rule-based baseline results}

Before applying our Machine Learning approach, we tested a baseline system based on syntactic and semantic rules to detect temporality from news. First, we created a financial semantic tree using the data provided by Multilingual Central Repository\footnote{A lexical database that integrates the Spanish WordNet into the EuroWordNet framework, available at {\tt http://adimen.si.ehu.es/web/MCR}, November 2020.} ({\sc mcr}) \citep{GonzalezEtAl12}. The hierarchical elements of interest were: ``commerce'', ``enterprise'', ``finance'', ``banking'', ``exchange'', ``money'', ``insurance'', ``tax'' and ``industry''. Then we applied the semantic tree to analyse the news content at word level. The analysis was divided into two parts:

\begin{itemize}
 \item We checked if any word in the news title was included in our financial semantic tree. Then, we created an extractive summary of the news content discarding those sentences that did not include any word belonging to the aforementioned financial semantic categories. The sentences that included the main ticker were kept. In the event that no word in the title was related to finance, based on the information extracted from the semantic tree, we created the extractive summary by applying a {\sc tf}-{\sc idf} approach with a relevance threshold of 0.75 inspired by similar works in the literature \citep{Rabelo2020,Suleman2021,Xiao2021}.
 
 \item By considering the dependency parsing analysis and using the extractive summary from the previous step, each sentence was analysed with Freeling. First, we considered the verbs with a syntactic relation with a ticker or a referential element acting as the main element of the clause. If no such verbs were detected, the closest previous or posterior verb to a ticker or referential element was taken into account. If no verbs were detected, we divided the sentences by commas and explored their parts separately. 
\end{itemize}

With a slight abuse of notation, let us refer to the number of instances of the respective verb tenses as ``past'', ``present'' and ``future''. We applied the following rules when either future or past $>$ 0 to decide the temporality of the news:

\begin{enumerate}
\item Future, if one of the following criteria is met:
\begin{enumerate}
 \item future $\geq$ past
 \item past$>$1 and present + future $>$ past
 \item present $\geq$ 3 $\times$ past
\end{enumerate}
\item Otherwise, past
\end{enumerate}

These rules have resulted from experimental tests with good performance. We performed a combinatorial search over configurable parameter ranges for the number of past, present and future references used as estimators.

Also based on experimental tests, where neither past nor future temporality was detected (when both future and past $=$ 0), the system considers that, if a present tense is followed by a number, it is a reference to the past (else, to the future). To explain the first condition take as an example the sentence ``On October 29, 2018, the stock dropped 6.6\%'' in Table \ref{tab:beforeafterpreproOriginal}, where the main verb is in past tense. In financial jargon this kind of statement is sometimes expressed in present tense. However, an explicit quantity is strongly indicative that the event has already occurred, unless there is an explicit reference to the future.

We obtained near-75\% accuracy using this rule-based baseline approach (see Table \ref{tab:rules}). We considered these results promising but they motivated us to pursue more sophisticated Machine Learning techniques to extract temporality from financial news content with higher performance.

\begin{table}[!htbp]
\centering
\caption{Macro performance of the rule-based baseline approach.}
\begin{tabular}{cccc}
\hline
\textbf{Precision} & \textbf{Recall} & {\bf Accuracy} \\ \toprule
73.48\% & 74.47\% & 74.57\%\\ \bottomrule
\end{tabular}
\label{tab:rules}
\end{table}

\subsection{Feature tuning and training}
\label{tuning_training}

To create the Machine Learning model, we chose the most adequate features from our annotated dataset.

For the generation of the $n$-grams (char-grams, word tokens and word-grams), we used {\it GridSearchCV}\footnote{Available at {\tt https://scikit-learn.org/stable/modules
/generated/sklearn.model\_selection.GridSearchCV.html}, November 2020.} from the Scikit-Learn Python library, which is an exhaustive search of an estimator in specified parameter ranges. We selected wide ranges in {\it CountVectorizer}\footnote{Available at {\tt https://scikit-learn.org/stable/modules/gene-
rated/sklearn.feature\_extraction.text.CountVectorizer.html}, November 2020.} (as shown in Listing \ref{configuration_parameters}). As a result, we obtained the following optimal parameters: {\tt max\_df} = 0.30, {\tt min\_df} = 0, {\tt ngram\_range} = (2,4) and {\tt max\_features} = 10000.

\begin{lstlisting}[frame=single,caption={Configuration parameters for the generation of $n$-grams.}, label={configuration_parameters}]
max_df: (0.3,0.35,0.4,0.5,0.7,0.8,1)
min_df: (0, 0.002, 0.005,0.008,0.01)
ngram_range: ((1, 1),(1, 2),(1, 3),(1, 4),(2, 4))
max_features: (10000,20000,30000,None)
\end{lstlisting}

\begin{table*}[!htbp]
\centering
\caption{Macro performance and training and testing times using selected textual features.}
\begin{tabular}{cccccc}
\hline
\textbf{Classifier} & \textbf{Precision} & \textbf{Recall} & {\bf Accuracy} & {\bf Train (s)} & {\bf Test (s)} \\ \toprule
ZeroR & 30.42 & 50.00 & 60.83 & $<0.01$ & $<0.01$ \\
DT & 65.82\% & 66.06\% & 67.17\% & 0.85 & 0.54 \\
RF & 77.48\% & 76.61\% & 78.17\% & 7.92 & 0.61\\
{\bf SVC} & {\bf 81.70\%} & {\bf 82.12\%} & {\bf 82.50\%} & 2.83 & 0.55\\
NN & 79.38\% & 79.33\% & 79.67\% & 380.46 & 0.89\\ \bottomrule
\end{tabular}
\label{tab:general}
\end{table*}

\begin{table*}[!htbp]
\centering
\caption{Macro performance and training and testing times using selected textual features plus all numerical and temporal features.}
\begin{tabular}{cccccc}
\hline
\textbf{Classifier} & \textbf{Precision} & \textbf{Recall} & {\bf Accuracy} & {\bf Train (s)} & {\bf Test (s)} \\ \toprule
ZeroR & 30.42 & 50.00 & 60.83 & $<0.01$ & $<0.01$ \\
DT & 67.12\% & 67.29\% & 68.33\% & 1.23 & 0.58\\
RF & 79.71\% & 78.59\% & 80.17\% & 5.81 & 0.64\\
{\bf SVC} & {\bf 83.08\%} & {\bf 82.38\%} & {\bf 83.50\%} & 3.60 & 0.60\\
NN & 82.68\% & 82.34\% & 83.17\% & 505.28 & 0.94\\ \bottomrule
\end{tabular}
\label{tab:generaltemporal}
\end{table*}

At the end, we obtained 30,000 features only for the $n$-grams (10,000 per type, char-grams, word-grams and word tokens). This huge set of features could lead to computationally unfeasible models and excessively long prediction times. Indeed, redundant features or poorly correlated features with the target predictor would cause unacceptable system performance. Therefore, we applied an attribute selector to extract the most relevant features with 10-fold cross-validation. 
For this purpose we chose the {\em SelectPercentile}\footnote{\label{note1}Available at {\tt https://scikit-learn.org/stable/modules/fea-
ture\_selection.html}, November 2020.} method from the Scikit-Learn Python library, as it outperformed other alternatives\footnoteref{note1} ({\em SelectFromModel}, {\em SelectKBest}, and {\em RFECV}). This method selects features according to a percentile of the highest scores. We set $Chi^2$ as the score function and an 80th percentile threshold. With this approach, $n$-gram features dropped to 24,000. 

Once these relevant $n$-gram features were selected, we analysed the temporal and numerical features, and we selected the best ones with a combinatorial analysis (one-vs-rest analysis for all 27 features) with 10-fold cross-validation (see Section \ref{machinelearning_results}). For all Machine Learning analyses, we applied a hyperparameter optimisation using {\it GridSearchCV} over the classification models with 10-fold cross-validation.

\subsection{Machine Learning results}
\label{machinelearning_results}

We performed three experiments: only using the selected $n$-grams (24,000 features), using the selected $n$-grams plus the two numerical features and all temporal features (25 features) and, finally, using the selected $n$-grams plus the most relevant numerical and temporal features from a combinatorial selection that we explain later in this section. In each of these three experiments we tested all Machine Learning classifiers in the system with 10-fold cross-validation on the experimental dataset. Since our main objective is detecting predictions of future events in news, we paid special attention to the results for this class.

Table \ref{tab:general} shows the results obtained with $n$-grams. The best classification model was {\sc svc} followed by {\sc nn}. However, the latter is computationally expensive due to the configuration time of the different hidden layers for each feature in the model. By comparing these results with those of the second experiment (Table \ref{tab:generaltemporal}), we observe a $\sim$4\% improvement in accuracy for the {\sc nn} classifier thanks to the numerical and temporal features. In any case, the improvement of the {\sc svc} classifier in Table \ref{tab:generaltemporal} over the rule-based baseline was at least 8\% for all metrics.

The effect of numerical and temporal features became more apparent when we checked the behaviour by class. Table \ref{tab:general_svc_nn} shows the results of the first experiment in that case. Note that precision and recall were very asymmetric between past and future ($\sim$10\% precision asymmetry with the {\sc svc} classifier, $\sim$19\% recall asymmetry with the {\sc nn} classifier). In addition, the precision of both classifiers was barely above 75\% for future.

\begin{table}[!htbp]
\centering
\caption{Performance by class using selected textual features.}
\begin{tabular}{cclcl}
\toprule
\multirow{2}{*}{\bf Classifier} & \multicolumn{2}{c}{\bf Precision} & \multicolumn{2}{c}{\bf Recall}\\
& Past & \multicolumn{1}{c}{Future} & Past & \multicolumn{1}{c}{Future}\\ \toprule
SVC & {\bf 87.12\%} & 76.28\% & 83.81\% & {\bf 80.42\%}\\ 
NN & 81.41\% & {\bf 77.21}\% & {\bf 86.28\%} & 67.63\%\\ 
\bottomrule
\end{tabular}
\label{tab:general_svc_nn}
\end{table}

\begin{table*}[!htbp]
\centering
\caption{Performance by class using selected textual features plus all numerical and temporal features.}
\begin{tabular}{cclcl}
\toprule
\multirow{2}{*}{\bf Classifier} & \multicolumn{2}{c}{\bf Precision} & \multicolumn{2}{c}{\bf Recall}\\
& Past & \multicolumn{1}{c}{Future} & Past & \multicolumn{1}{c}{Future}\\ \toprule
SVC & 85.89\% & {\bf 80.28\%} & {\bf 87.36\%} & 77.39\%\\ 
NN & {\bf 86.47\%} & 78.88\% & 85.98\% & {\bf 78.71\%}\\
\bottomrule
\end{tabular}
\label{tab:generaltemporal_svc_nn}
\end{table*}

The introduction of temporal features solves these performance issues (asymmetric precision and levels under 80\%) to a great extent. Accordingly, Table \ref{tab:generaltemporal_svc_nn} shows a 4\% improvement in precision for the future class with a {\sc svc} classifier. For the {\sc nn} classifier, precision improvements were $\sim$5\% and $\sim$2\% for past and future, respectively, although the precision for the future class only reached $\sim79$\%. Note however the high improvement in recall with the {\sc nn} classifier for the future class, which exceeded 10\%.

We then focused on the {\sc svc} classifier, whose recall for the future class was still under $80$\% but presented the best overall results in terms of precision, recall and accuracy. Here we introduced the previously mentioned selection of the most relevant numerical and temporal features with a combinatorial analysis with 10-fold cross-validation. We chose 11 such features due to the precision they attained: number of present and future verbs from the dependency analysis with the ticker acting as subject, number of present verbs from the dependency parsing with the ticker acting either as subject or object, number of present and future verbs from the proximity analysis with the ticker acting as subject, number of present and future verbs from the proximity analysis with the ticker acting either as subject or object, global temporality from the proximity analysis with the ticker acting either as subject or object, number of future verbs in the second part of the news piece and number of present and future verbs in the third part. Only the features that produced precision symmetry between classes were kept, and both numerical features were discarded.

\begin{table*}[!htbp]
\centering
\caption{Macro performance and training and testing times using selected textual features and most relevant temporal features from the combinatorial analysis.}
\begin{tabular}{cccccc}
\hline
\textbf{Classifier} & \textbf{Precision} & \textbf{Recall} & {\bf Accuracy} & {\bf Train (s)} & {\bf Test (s)} \\ \toprule
{\bf SVC} & {\bf 83.88\%} & {\bf 83.59\%} & {\bf 84.33\%} & 3.47 & 0.60\\
NN & 79.80\% & 79.22\% & 80.00\% & 571.25 & 0.94\\ \bottomrule
\end{tabular}
\label{tab:total}
\end{table*}

Table \ref{tab:total} shows that, with this second selection, we attained well over 80\% precision and recall performance with the {\sc svc} classifier, which takes considerably less time to train than the {\sc nn}. Furthermore, Table \ref{tab:generalrelevant_temporal_svc_nn} shows the precision and recall of the {\sc svc} classifier by class. Note that all metrics exceeded 80\% as pursued, a level that, compared to other Machine Learning financial applications in the literature \citep{Zhu2017,Atkins2018,Zhu2019,Dridi2019,DeArriba-Perez2020}, is similar and even superior.

\begin{table}[!htbp]
\centering
\caption{Performance by class using selected textual features and most relevant temporal features from the combinatorial analysis, {\sc svc} classifier.}
\begin{tabular}{cclcl}
\toprule
\multirow{2}{*}{\bf Classifier} & \multicolumn{2}{c}{\bf Precision} & \multicolumn{2}{c}{\bf Recall}\\
& Past & \multicolumn{1}{c}{Future} & Past & \multicolumn{1}{c}{Future}\\ \toprule
SVC & 87.54\% & 80.21\% & 86.79\% & 80.40\%\\ 
\bottomrule
\end{tabular}
\label{tab:generalrelevant_temporal_svc_nn}
\end{table}

\subsection{Application use case}

\begin{figure}[!htbp]
\centering
\includegraphics[scale=0.15]{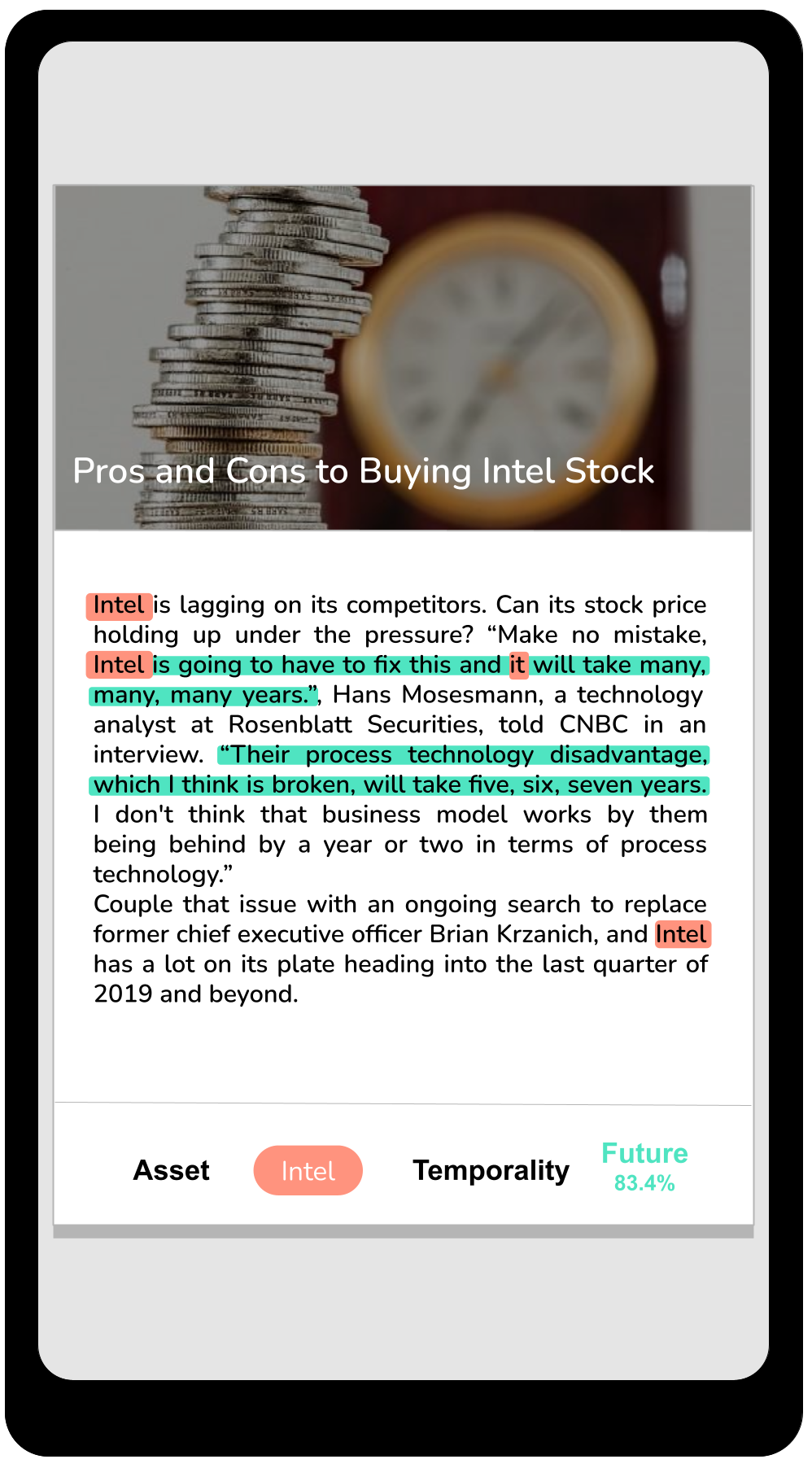}
\caption{\label{use_case} Example of temporality detection system for financial news integrated into a mobile application.}
\end{figure}

A direct use case of our system would be a financial mobile application (``app'') for investors, which would automatically process news sources. The app would translate this vast amount of public information into key indicators on assets, including overall temporality of key statements. Figure {\ref{use_case}} shows a possible interface of this app, where future statements are marked in green and assets are highlighted in pink. The confidence in the classification is also presented to the users at the bottom.

\section{Conclusions}
\label{sec:conclusions}

Motivated by the amount of publicly available data about stock markets in financial news, we propose a novel system to detect the temporal dimension of the key statements in a text at discourse level, which relies on a combination of {\sc nlp} and Machine Learning techniques.

For that purpose, we departed from a manually annotated dataset of 600 financial news. In addition to textual features ($n$-grams) and numerical features, we applied different analyses to the text to obtain temporal features: dependency parsing, verb extraction by proximity and verb analysis by text partitions (as an approximation to discourse patterns in financial news). Thus, the more sophisticated features required syntactic and dependency parsing {\sc nlp} techniques. These features were essential for the Machine Learning algorithms to achieve precision and recall above $80$\%, both globally and by class.

Since we are not aware of any previous research on this same problem, we checked our approach against a rule-based baseline based on the temporal features. The Machine Learning approach was significantly better (between 8\% and 10\% improvement in all metrics).

As future work we plan to design a multilingual version of our architecture to also cover financial news in Spanish.

\section*{Acknowledgements}\label{sec:acknol}

This work was partially supported by Xunta de Galicia grants GRC2018/053 and ED341D-R2016/012.

\bibliography{bibliography}

\end{document}